\documentclass[10pt,twocolumn,letterpaper]{article}

\usepackage{cvpr}
\usepackage{times}
\usepackage{epsfig}
\usepackage{graphicx}
\usepackage{amsmath}
\usepackage{amssymb}
\usepackage{multirow}
\usepackage{booktabs}
\usepackage{cite}

\usepackage[table]{xcolor}
\usepackage{colortbl}
\usepackage{float}

\usepackage{booktabs}
\usepackage{gensymb}
\usepackage{multirow}
\usepackage[norule,symbol,perpage]{footmisc}
\usepackage[caption=false,font=footnotesize]{subfig}

\newcommand{\app}{\raise.17ex\hbox{$\scriptstyle\sim$}}

\def\pt{p_\textrm{t}}
\def\qt{q_\textrm{t}}
\def\rt{r_\textrm{t}}

\def\at{\alpha_\textrm{t}}

\def\CE{\textrm{CE}}
\def\FL{\textrm{FL}}

\def\CMFL{\textrm{\textit{CMFL }}}

\def\cmfl{\textrm{\textit{CMFL}}}

\def\hqwmca{\textrm{\textit{HQ-WMCA }}}
\def\wmca{\textrm{\textit{WMCA }}}

\usepackage[pagebackref=true,breaklinks=true,letterpaper=true,colorlinks,bookmarks=false]{hyperref}

\cvprfinalcopy 

\ifcvprfinal\fi
\begin{document}

\title{Cross Modal Focal Loss for RGBD Face Anti-Spoofing} 

\author{Anjith George and S\'ebastien Marcel \\
Idiap Research Institute \\
Rue Marconi 19, CH - 1920, Martigny, Switzerland \\
{\tt\small  \{anjith.george, sebastien.marcel\}@idiap.ch  }
}

\maketitle

\begin{abstract}

Automatic methods for detecting presentation attacks are essential to ensure the reliable use of facial recognition technology. Most of the methods available in the literature for presentation attack detection (PAD) fails in generalizing to unseen attacks. In recent years, multi-channel methods have been proposed to improve the robustness of PAD systems. Often, only a limited amount of data is available for additional channels, which limits the effectiveness of these methods. In this work, we present a new framework for PAD that uses RGB and depth channels together with a novel loss function. The new architecture uses complementary information from the two modalities while reducing the impact of overfitting. Essentially, a cross-modal focal loss function is proposed to modulate the loss contribution of each channel as a function of the confidence of individual channels. Extensive evaluations in two publicly available datasets demonstrate the effectiveness of the proposed approach.

\end{abstract}

\section{Introduction}

While face recognition technology has become a ubiquitous method for biometric authentication, the vulnerability to presentation attacks (also known as ``spoofing attacks'') is a major concern when used in secure scenarios \cite{costa2016replay}, \cite{erdogmus2014spoofing}. These attacks can be either \textit{impersonation} or \textit{obfuscation} attacks. Impersonation attacks attempt to gain access by masquerading as someone else and obfuscation attacks attempt to evade face recognition systems. While many methods have been suggested in the literature to address this problem, most of these methods fail in generalizing to unseen attacks \cite{de2013can}. In a practical scenario, it is not possible to anticipate all the types of attacks at the time of training a PAD model. Moreover, a PAD system is expected to detect new types of sophisticated attacks. It is therefore important to have unseen attack robustness in PAD models.

\begin{figure}[t]
\centering
        \includegraphics[width=0.98\linewidth]{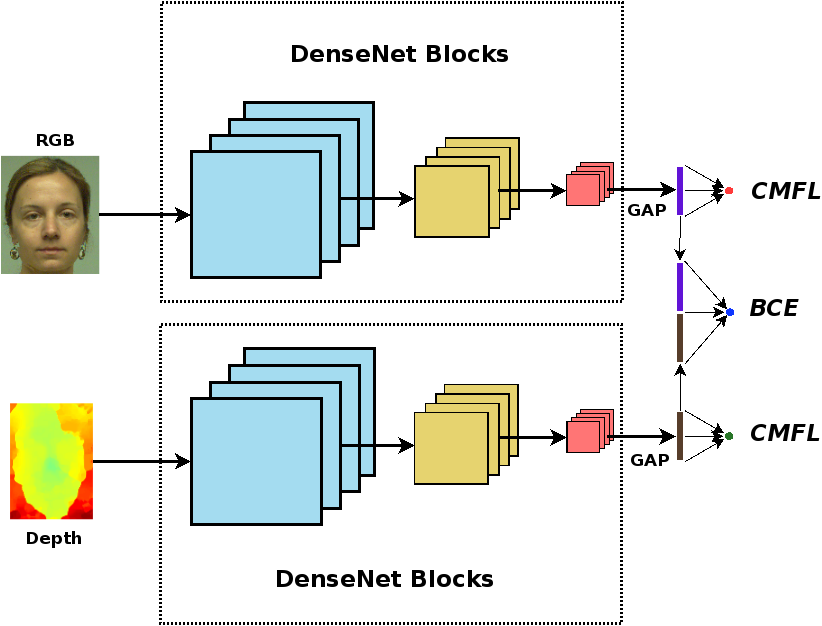}
        \caption{The proposed framework for PAD. A two stream- multi-head architecture is used following a late fusion strategy. Heads corresponding to individual channels are supervised by the proposed cross-modal focal loss (CMFL), while the joint model is supervised by binary cross entropy (BCE).}
        \label{fig:framework}

\end{figure} 

The majority of the literature deals with the detection of these attacks with RGB cameras. Over the years, many feature-based methods have been proposed using color, texture, motion, liveliness cues, histogram features \cite{boulkenafet2015face}, local binary pattern \cite{maatta2011face}, \cite{chingovska2012effectiveness} and motion patterns \cite{anjos2011counter} for performing PAD. Recently several CNN based methods have also been proposed including 3D-CNN \cite{gan20173d}, part-based models \cite{li2017face} and so on. Some works have shown that using auxiliary information in the form of binary or depth supervision improves performance \cite{atoum2017face,george2019deep}. However, most of these methods have been designed specifically for 2D attacks and the performance of these methods against challenging 3D and partial attacks is poor \cite{liu2019deep}. Moreover, these methods suffer from poor unseen attack robustness.

The performance of RGB only models deteriorates with sophisticated attacks such as 3D masks and partial attacks. Due to the limitations of visible spectrum alone, several multi-channel methods have been proposed in literature such as \cite{raghavendra2017extended}, \cite{erdogmus2014spoofing}, \cite{steiner2016reliable}, \cite{dhamecha2013disguise}, \cite{agarwal2017face}, \cite{Bhattacharjee:256262}, \cite{bhattacharjee2017you}, \cite{george_mccnn_tifs2019,heusch2020deep,george2020face,george_mcocm_tifs2020} for face PAD. Essentially, it becomes more difficult to fool a multi-channel PAD system as it captures complementary information from different channels.  Deceiving different channels at the same time requires considerable effort. Multi-channel methods have proven to be effective, but this comes at the expense of customized and expensive hardware. This could make these systems difficult to deploy widely, even if they are robust. A variety of channels are available for PAD, e.g., RGB, depth, thermal, NIR spectra, SWIR spectra, ultraviolet, light field imagery, etc. Out of these various modalities, we find that RGB-D devices are commercially available and are quite affordable, making it possible to deploy them in real-world scenarios. The  Intel RealSense family of devices, Microsoft Kinect, and the OpenCV AI Kit (OAK-D) \cite{oak_d} are examples of standard devices that do not require any additional effort to obtain multi-channel images. Due to the wide availability of these channels in an integrated package, we choose RGB and Depth as the two channels to be used in this work. However, the proposed framework can be trivially extended to any combinations of channels.

Even when using multiple channels, the models tend to overfit to attacks seen in the training set. While the models could perform perfectly in attacks seen in the training set, degradation in performance is often observed when confronted with unseen attacks in real-world scenarios. This is a common phenomenon with most of the machine learning algorithms, and this problem is aggravated in case of a limited amount of training data. The models, in the lack of strong priors, could overfit to the statistical biases of specific datasets it was trained on and could fail in generalizing to unseen samples. Multi-channel methods also suffer from an increased possibility of overfitting as they increase the number of parameters due to the extra channels. 

In this work, we address this issue in two different directions. First, we use a multi-head architecture, which follows a late fusion strategy to combine different channels of information. Instead of joining the representations into a joint final node, we keep three different heads separately for the individual branches and the joint branch, this can be viewed as a form of architectural regularization. The proposed architecture can be found in Fig. \ref{fig:framework}. This enables us a way to supervise individual channels together with the joint representation, ensuring robust representation would be learned in individual as well as joint branches. Secondly, we propose a cross-modal focal loss function to supervise the individual channels that modulate the loss function factoring in the confidence of the channels present. 

The main contributions of this work are listed below:

\begin{itemize}

\item A frame-level RGB-D face PAD method is proposed which operates on synchronously captured RGB-D samples.

\item A new loss function called cross-modal focal loss (\CMFL) is proposed, which can be used to supervise individual channels in a multi-stream architecture.

\item Though the model is trained for a multi-channel scenario, it can also be deployed with individual channels by just using the score from the head corresponding to the available channel.

\item We show the efficacy of the proposed framework in two publicly available datasets consisting of a wide variety of challenging unseen attacks.

\end{itemize}

Additionally, the source code and protocols to reproduce the results are available publicly\footnote{Source code: \url{will-be-available-upon-acceptance}}.

\section{Proposed approach}

Different stages of the proposed PAD framework are described in this section. 

\subsection{Preprocessing}

\label{subsec:preprocess}

The PAD pipeline acts on the cropped facial images. For the RGB image, the preprocessing stage consists of face detection and landmark localization using the MTCNN \cite{zhang2016joint} framework, followed by alignment. The detected face is aligned by making the eye centers horizontal followed by resizing them to a resolution of $224 \times 224$. For the depth image, a normalization method using the median absolute deviation (MAD) \cite{nikisins2019domain} is used to normalize the face image to an 8-bit range. The raw images from RGB and depth channels are already spatially registered so that the same transformation can be used to align the face in the depth image.

\subsection{Network architecture and loss function}

This section details the proposed network architecture and loss function. 

\subsubsection{Architecture}

From the prevailing literature, it has been observed that multi-channel methods are robust against a wide range of attacks \cite{george_mccnn_tifs2019,heusch2020deep,george2020face,george_mcocm_tifs2020}. Broadly, there are four different strategies to fuse the information from multiple channels, they are 1) early fusion, meaning the channels are stacked at the input level (for example, MC-PixBiS \cite{heusch2020deep}). The second strategy is late fusion, meaning the representations from different networks are combined at a later stage similar to feature fusion (for example MCCNN \cite{george_mccnn_tifs2019}), a third strategy is score level fusion where individual networks are trained separately for different channels and score level fusion is performed on the scalar scores from each channel. A fourth strategy is a hybrid approach where information from multiple levels is combined as in \cite{parkin2019recognizing}. 

Though multiple channels can perform well with a wide variety of attacks, they tend to overfit to known attacks when all channels are used together and trained as a binary classifier. To avoid this, we propose a multi-head architecture that follows a late fusion strategy. The architecture of the proposed network is shown in Fig. \ref{fig:framework}. Essentially, the architecture consists of a two-stream network with separate branches for the component (RGB and Depth) channels. The embeddings from the two channels are combined to form the third branch. Fully connected layers are added to each of these branches to form the final classification head. These three heads are jointly supervised by a loss function which forces the network to learn discriminative information from individual channels as well as the joint representation, reducing the possibility of overfitting. The multi-head structure also makes it possible to perform scoring even when a channel is missing at test time, meaning that we can do scoring with RGB branch alone (just using the score from the RGB head) even if the network was trained on RGB-D. 

The branches are comprised of the first eight blocks from DenseNet architecture (densenet161) proposed by Huang \textit{et al}. \cite{huang2017densely}. In the DenseNet architecture, each layer is connected to every other layer, reducing the vanishing gradient problem while reducing the number of parameters. We used pre-trained weights from the Image Net dataset to initialize the individual branches. The number of input channels for the RGB and depth channels has been modified to 3 and 1 for the RGB and depth channels, respectively. For the depth branch, the mean values of three-channel weights are used to initialize the weights of the modified convolutional kernels in the first layer. In each branch, a global average pooling (GAP) layer is added after the dense layers to obtain a 384-dimensional embedding. The RGB and depth embeddings are concatenated to form the joint embedding layer. A fully connected layer, followed by a sigmoid activation is added on top of each of these embeddings to form the different heads in the framework. At training time, each of these heads is supervised by a separate loss function. At test time, the score from the RGB-D branch is used as the PAD score.

\subsubsection{Cross Modal Focal Loss function (\CMFL)}

Having individual heads make it possible to train a multi-channel model with the capability to handle missing channels at test time. Now, one naive way to supervise this network would be to supervise the individual branches with binary cross-entropy (BCE).   

However, the usage of BCE for individual channels may not be ideal. The issue is illustrated as follows; we can consider different channels as different views of the same sample, and for some attacks, it may not be possible to distinguish it just from one view. It is possible that the images of some attacks would look perfectly like bonafide samples when viewed in just one channel. For example, facial makeup when viewed in depth channel would look exactly like the depth map of a bonafide sample. The naive way of supervising the depth branch with BCE in such cases might lead to overfitting. However, in the same scenario, discriminating makeup would be more obvious in RGB and joint representations. From this example,  it can be seen that supervising the individual branches separately might not result in robust decision boundaries. One way to approach this issue is to use the prediction probabilities from the \textit{current} branch and the \textit{other} branch to change the loss contributions of samples in each branch. We propose a cross-modal focal loss function to supervise the individual channels, which modulates the loss function based on the confidence of the current and alternate channel present. 

For each branch, the samples which could be classified correctly should be well separated in the score space. At the same time, we encourage the individual branches to produce unsure scores when there is not enough discriminatory information, rather than overfitting to some statistical bias in the training data. However, this applies only when the other branch can confidently classify the sample correctly. 

More formally, consider a binary classification problem where the samples are multi-modal i.e., each sample is a pair of images or features that capture different views with complementary information. Now assume that the classification problem cannot be done with a single channel alone (or is a very hard problem). Combining the features from both the channels and using a learning strategy using the joint features could provide a solution for this. However, this could lead to overfitting and cannot handle missing channels at test time.

\begin{figure}[t]
\centering
        \includegraphics[width=0.8\linewidth]{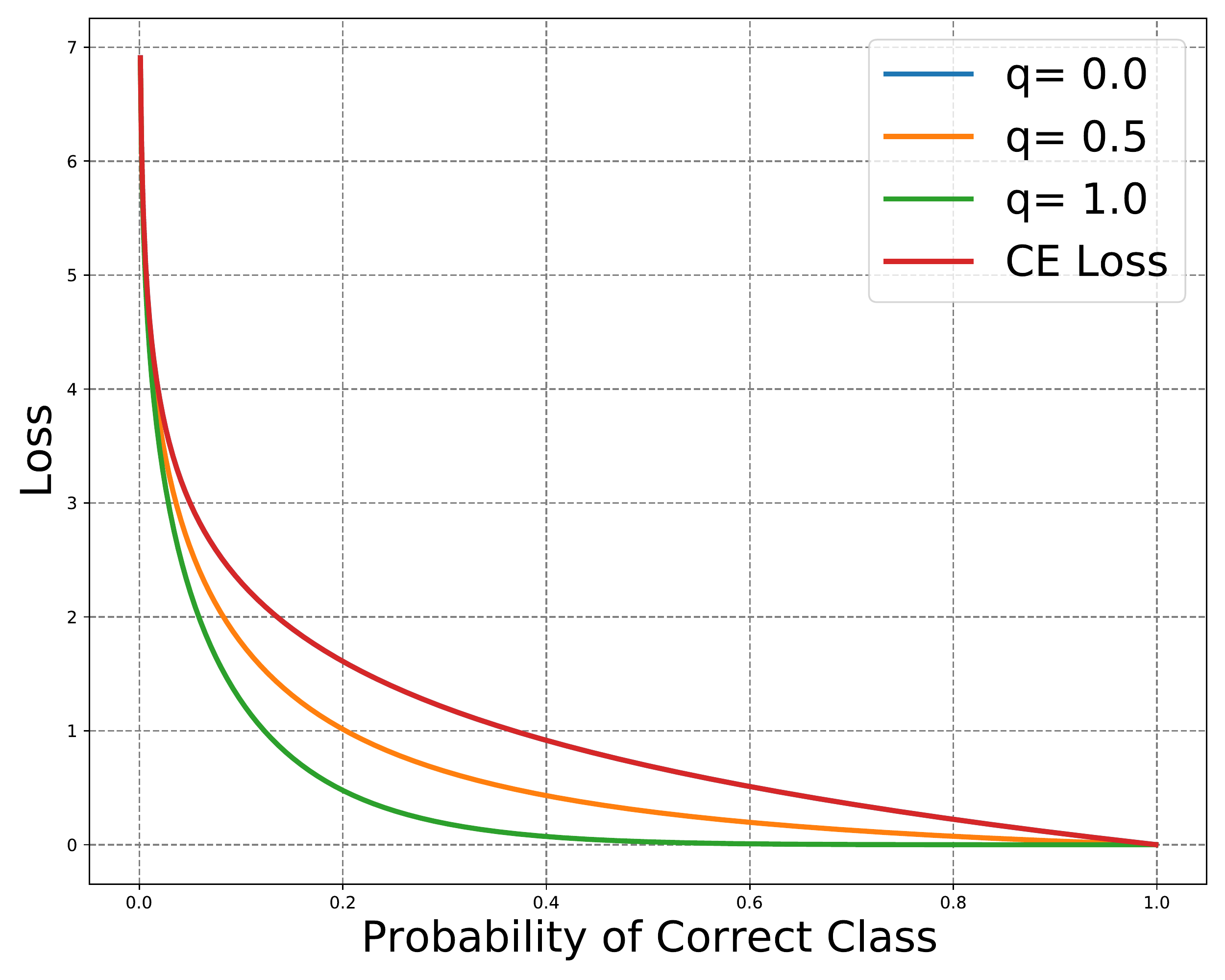}
        \caption{The loss curve for the proposed loss function ($\gamma =3$),  the variable $q$ represents the probability of correct classification from the other branch in the two-stream network. It can be seen that the loss curve converges to cross entropy loss when $q=0$. The loss contribution reduces as the value of $q \rightarrow 1$.}
\label{fig:losscurve}
\end{figure}

If we use BCE on individual branches, the loss will penalize heavily the samples which cannot be classified with the information available with the specific channel. In such a scenario, the model might start to overfit to the biases in the dataset to minimize the loss function, resulting in an over-fitted model.

To avoid this, we propose cross-modal focal loss (CMFL) to supervise the individual channels. The core idea is that, when one of the channels can correctly classify a sample correctly with high confidence, then the loss contribution of the sample in the other branch can be reduced. If a channel can correctly classify a sample confidently, then we don't want the other branch to penalize the model more. CMFL forces each branch to learn robust representations for individual channels, which can then be utilized with the joint branch, effectively acting as an auxiliary loss function.

The idea of relaxing the loss contribution of samples correctly classified is similar to the Focal Loss \cite{lin2018focal} used in object detection problems. In Focal Loss, a modulating factor is used to reduce the loss contributed by samples that are correctly classified with high confidence. We use this idea by modulating the loss factoring in the confidence of the sample in the current and the alternate branch. 

Consider the expression for cross-entropy (CE) in a binary classification problem:

\begin{equation}
\CE(p,y) = \begin{cases} -\log(p) &\text{if $y = 1$}\\
 -\log (1 - p) &\text{if $y = 0$}\end{cases}
\end{equation}

Where $y \in \{ 0,1\}$  denotes the class label (y:0 attack, y:1 bonafide) and $p \in [0,1]$ is the probability of the class. We follow a similar notation $\pt$ as in \cite{lin2018focal}, which is the probability of the target class:

\begin{equation}
\pt=\begin{cases} p &\text{if $y = 1$}\\
1 - p &\text{otherwise,}\end{cases}
\end{equation}

and write $\CE(p,y) = \CE(\pt) = - \log (\pt)$.

In the $\alpha$-balanced form CE loss can be written as:
\begin{equation}
\CE(\pt) = - \at \log (\pt).
\end{equation}

The standard $\alpha$-balanced focal loss (FL) \cite{lin2018focal} adds a modulating factor $(1 - \pt)^\gamma$ to the cross entropy loss, with a tunable \emph{focusing} parameter $\gamma \ge 0$, making the loss formulation as. 
\begin{equation}
\FL(\pt) = - \at (1 - \pt)^\gamma \log (\pt).
\end{equation}

\begin{figure}[h!]
\centering
        \includegraphics[width=0.6\linewidth]{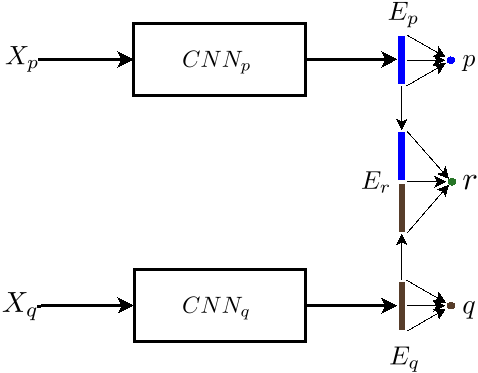}
        \caption{Diagram of the two-stream multi-head modal, showing the embeddings and probabilities from individual and joint branches. This can be extended to multiple heads as well.}
\label{fig:model}
\end{figure} 

Consider the two-stream multi-branch multi-head modal in Fig. \ref{fig:model}. $X_p$ and $X_q$ denotes the image inputs from different
modalities, and $E_p$, $E_q$, and $E_r$ denotes the corresponding embeddings for the individual and joint representations. In each branch, after the embedding layer, a fully connected layer (followed by a sigmoid layer) is present which provides classification probability. The variables $p$, $q$ and $r$ denote
these probabilities. 

The proposed Cross Modal Loss Function (CMFL) is given as follows:

\begin{equation}
\cmfl{\pt,\qt} = - \at (1 - w(p_t,q_t))^\gamma \log (\pt)
\end{equation}

The function $w(p_t,q_t)$, depends on the probabilities given by the channels from two individual branches. 
This modulating factor should increase as the probability of the other branch increases, and at the same time
should be able to prevent very confident mistakes. Hence for this study, we use the harmonic mean of both the branches
weighted by the probability of the other branch. This reduces the loss contribution when the other branch is giving confident predictions. 
And the expression for this function is given as:

\begin{equation}
w(p_t,q_t) = q_{t}\frac{2p_{t}q_{t}}{p_{t}+q_{t}}
\end{equation}

Note that the function $w$ is assymetric, i.e., the expression for $w(q_t,p_t)$ is:

\begin{equation}
w(q_t,p_t) = p_{t}\frac{2p_{t}q_{t}}{p_{t}+q_{t}}
\end{equation}
Meaning the weight function depends on the probability of the other branch. Now we use the proposed loss function as auxiliary supervision,and the overall loss 
function to minimize is given as:

\begin{equation}
\mathcal{L}=(1-\lambda) \mathcal{L}_{\CE(\rt)} + \lambda (\mathcal{L}_{\cmfl{\pt,\qt}} + \mathcal{L}_{\cmfl{\qt,\pt}})
\end{equation}

We have set the value of $\lambda$ non-optimally as $0.5$ for this study. The loss curve for the cross-entropy and the proposed loss is shown in Fig. \ref{fig:losscurve}. 
When the probability of other branch is zero, then the loss is equivalent to standard cross-entropy. The loss contribution is reduced when the other branch is able to correctly classify the
sample. ie, when an attack example is misclassified by network $CNN_p$, the network $CNN_p$ is penalized unless model $CNN_q$ can classify the attack sample with high confidence. As the $w(p,q) \rightarrow 1$ the modulating factor goes to zero, meaning if one channel is able to classify it perfectly, then the other branch is less penalized. Also, the focussing parameter $\gamma$ can be adapted to change the behaviour of the loss curve. We used an empirically obtained value of $\gamma = 3$  in all our experiments.

Without the loss of generality, the framework can be extended to other multi-channel classification problems as well.

\subsubsection{Implementation details}

We performed data augmentation during the training phase with random horizontal flips with a probability of 0.5. The combined loss function is minimized with Adam Optimizer \cite{kingma2014adam}. A learning rate of $1\times10^{-4}$ was used with a weight decay parameter of  $1\times10^{-5}$. We used a mini-batch size of 64, and the network was trained for 25 epochs on a GPU grid. During the evaluation of the model, the scores from the RGB-D head was used to calculate the final PAD score. The proposed architecture has about 6.39M parameters and about 9.16 GFLOPS. The architecture and the training framework were implemented using the PyTorch \cite{paszke2017automatic} library.

\section{Experiments and Results}


\begin{figure*}[ht]
\centering
  \subfloat[]{\includegraphics[height=2.3cm]{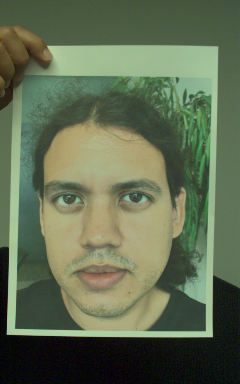}}%
\hfil
  \subfloat[]{\includegraphics[height=2.3cm]{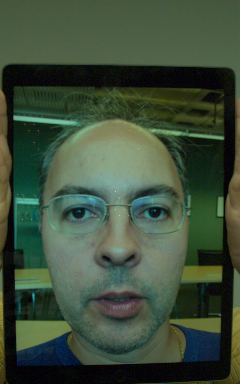}}%
\hfil
  \subfloat[]{\includegraphics[height=2.3cm]{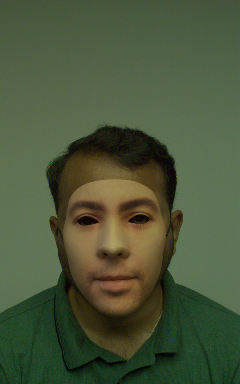}}%
\hfil
  \subfloat[]{\includegraphics[height=2.3cm]{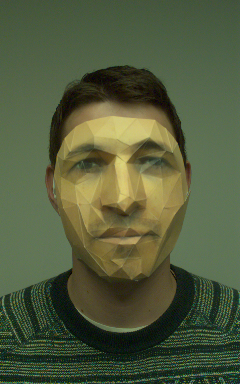}}%
\hfil
  \subfloat[]{\includegraphics[height=2.3cm]{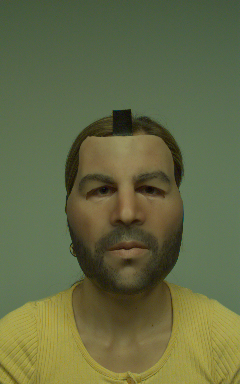}}%
\hfil
  \subfloat[]{\includegraphics[height=2.3cm]{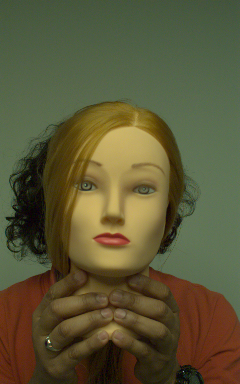}}%
\hfil
  \subfloat[]{\includegraphics[height=2.3cm]{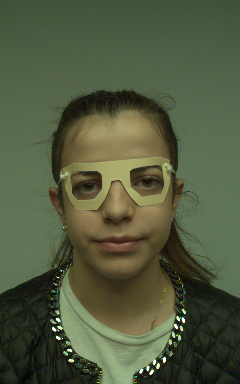}}%
\hfil
  \subfloat[]{\includegraphics[height=2.3cm]{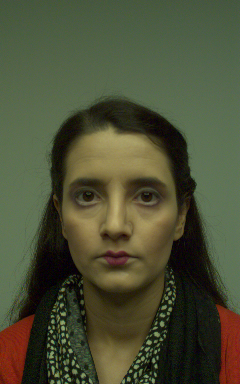}}%
\hfil
  \subfloat[]{\includegraphics[height=2.3cm]{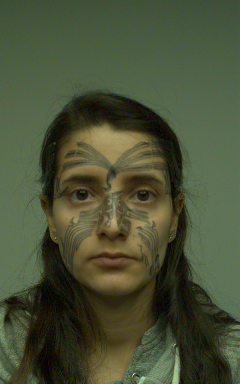}}%
\hfil
  \subfloat[]{\includegraphics[height=2.3cm]{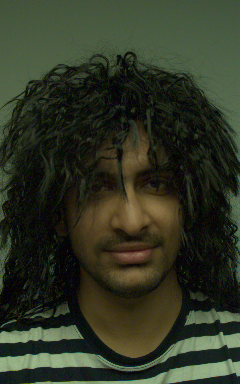}}%
  \caption{Attacks present in \hqwmca dataset: (a) Print, (b) Replay, (c) Rigid mask, (d) Paper mask, (e) Flexible mask, (f) Mannequin,
          (g) Glasses, (h) Makeup, (i) Tattoo and (j) Wig. Image taken from \cite{heusch2020deep}.}
\label{fig:attacks}
\end{figure*}

\subsection{Datasets used}
We have used two publicly available datasets for the experiments, namely \wmca and \hqwmca, which contains a wide variety of 2D, 3D, and partial attacks.
\subsubsection{\wmca dataset}
The \textit{Wide Multi-Channel presentation Attack} (\wmca) \cite{george_mccnn_tifs2019} database contains a wide variety of 2D and 3D presentation attacks, with  a total of \textit{1679} video samples from \textit{72} subjects.  Multiple channels collected synchronously, namely color, depth, infrared, and thermal channels, collected using two consumer-grade devices,  Intel\textsuperscript{\textregistered} RealSense\texttrademark SR300 (for color, depth and infrared), and Seek Thermal CompactPRO (for the thermal channel) is available with this database.  Though four different channels are available in this database, in this work,  we focus on the RGB and depth data obtained from the Intel\textsuperscript{\textregistered} RealSense\texttrademark SR300 device.
\subsubsection{\hqwmca dataset}
The High-Quality Wide Multi-Channel Attack (\hqwmca)\cite{heusch2020deep,Mostaani_Idiap-RR-22-2020} dataset consists of 2904 short multi-modal video recordings of both bonafide and presentation attacks. This database again consists of a wide variety of attacks including both obfuscation and impersonation attacks. Specifically, the attacks considered are print, replay, rigid mask, paper mask, flexible mask, mannequin, glasses, makeup, tattoo, and wig (Fig. \ref{fig:attacks}). The database consists of recordings from 51 different subjects, with several channels including color, depth, thermal, infrared (spectra), and short-wave infrared (spectra). In this work, we consider the RGB channel captured with Basler acA1921-150uc camera and depth image captured with Intel RealSense D415.
\subsection{Protocols}
Since both of the datasets contains a wide variety of attacks, we have created leave-one-out (LOO) attack protocols individually for both of the datasets. Specifically, one attack is left out in the train and development set and the evaluation set consists of bonafide and the attack which was left out in the train and development set. This constitutes the unseen attack protocols or zero-shot attack protocols. The performance of the PAD methods in these protocols gives a more realistic estimate of their robustness against unseen attacks in real-world scenarios. Further, for cross-dataset experiments, we have created \textit{grandtest} protocols in both of the datasets which consist of attacks distributed in the train, development, and test sets (with disjoint identities across folds).
\subsection{Metrics}
For the evaluation of the algorithms, we have used the ISO/IEC 30107-3 metrics \cite{ISO}, Attack Presentation Classification Error Rate (APCER), and Bonafide Presentation Classification Error Rate (BPCER) along with the Average Classification Error Rate (ACER) in the $eval$ set. We compute the threshold in the $dev$ set for a BPCER value of 1\%.
\begin{equation}
ACER=\frac{APCER+BPCER}{2}.
\end{equation}
For cross-database testing, Half Total Error Rate (HTER) is adopted following the convention in \cite{george_mcocm_tifs2020}, which computes the average of False Rejection Rate (FRR) and the False Acceptance Rate (FAR). HTER is computed in the $eval$ set using the threshold computed in the $dev$ set using the equal error rate criterion (EER).  
\subsection{Baseline methods}
For a fair comparison with state of the art, we have implemented 3 different multi-channel PAD approaches from literature for the RGB-D channels. Besides, we also introduce the proposed multi-head architecture supervised with BCE alone, as another baseline for comparison. The baselines implemented are listed below.\\ \textbf{\textit{MC-PixBiS}}:  This is a CNN based system \cite{george2019deep}, extended to multi-channel scenario as described in \cite{heusch2020deep} trained using both binary and pixel-wise binary loss function. This model uses RGB and depth channels stacked together at the input level.\\
\textbf{\textit{MCCNN-OCCL-GMM}}: This model is the multi-channel CNN system proposed to learn one class model using the one class contrastive loss (OCCL) and Gaussian mixture model as reported in \cite{george_mcocm_tifs2020}. The model was adapted to accept RGB-D channels as the input.\\
 \textbf{\textit{MC-ResNetDLAS}}: This is the reimplementation of the architecture from \cite{parkin2019recognizing}, which won the first prize in the `CASIA-SURF' challenge, extending it to RGB-D channels, based on the open-source implementation \footnote{Available from: \url{https://github.com/AlexanderParkin/ChaLearn_liveness_challenge}}. We used the initialization from the best-pretrained model as suggested in \cite{parkin2019recognizing} followed by retraining in the current protocols using RGB-D channels.\\
 \textbf{\textit{RGBD-MH-BCE}}: This uses the newly proposed multi-head architecture shown in Fig.\ref{fig:framework}, where all the branches are supervised by binary cross-entropy (BCE). In essence, this is equivalent to setting the value of $\gamma = 0$, in the expression for the cross-modal loss function. This is shown as a baseline to showcase the improvement by the new multi-head architecture alone and to contrast with the performance change with the new loss function. \\
 \textbf{\textit{Proposed}}: This is our final proposed framework, it uses the multi-head architecture we proposed as shown in Fig.\ref{fig:framework}, together with the newly proposed loss function. More specifically, the individual channel branches are supervised by the newly proposed cross-modal focal loss function (\CMFL). \\
The details of the parameters of the baseline methods can be found in our open-source implementation \footnote{Will be available upon acceptance.}.

\subsection{Experiments}
We have conducted experiments in both \wmca and \hqwmca datasets, specifically the leave-one-out protocols to evaluate the robustness against unseen attacks. The results are described in the following sections.
\subsubsection{Results in \wmca dataset}
The performance of the proposed system and baselines in the LOO protocols of \wmca are shown in Table \ref{tab:unseen_wmca}. It can be seen that the baselines \textit{MCCNN-OCCL-GMM} and \textit{MC-ResNetDLAS} performs poorly in the challenging unseen attack scenarios. Among baselines, the \textit{MC-PixBiS} model achieves the best performance with an average ACER of $10.5 \pm 16.7 \%$. The model with the new architecture, \textit{RGBD-MH-BCE} achieves reasonable performance as compared to the baselines. However, the proposed method, with the new \CMFL loss achieves the best mean accuracy. Also, comparing with \textit{RGBD-MH-BCE}, the proposed loss function shows a clear improvement in performance achieving an average ACER of $7.6 \pm 11.2 \%$. 
\begin{table*}[ht]
\centering
\caption{Performance of the baseline systems and the proposed method in \textbf{unseen} protocols of \wmca dataset. The values reported are obtained with a threshold computed for BPCER 1\% in $dev$ set.}
\label{tab:unseen_wmca}

\resizebox{0.95\textwidth}{!}{%

\begin{tabular}{lrrrrrrr>{\columncolor[gray]{0.8}}r}
\toprule
{}                                      &  Flexiblemask &  Replay &  Fakehead &  Prints &  Glasses &  Papermask &  Rigidmask &  Mean$\pm$Std \\
\midrule
MC-PixBiS \cite{george2019deep}                               &          49.7 &     3.7 &       0.7 &     0.1 &     16.0 &        0.2 &        3.4 &  10.5$\pm$16.7 \\
MCCNN-OCCL-GMM \cite{george_mcocm_tifs2020}                         &          22.8 &    31.4 &       1.9 &    30.0 &     50.0 &        4.8 &       18.3 &  22.7$\pm$15.3 \\
MC-ResNetDLAS \cite{parkin2019recognizing}                           &          33.3 &    38.5 &      49.6 &     3.8 &     41.0 &       47.0 &       20.6 &  33.4$\pm$14.9 \\ \hline
RGBD-MH-BCE       &            33.7 &    1.0 &      3.1 &    1.7 &    37.6 &        1.0 &        2.2 &  11.4$\pm$15.3 \\
\textbf{Proposed} &            12.4 &    1.0 &      2.5 &    0.7 &    33.5 &        1.8 &        1.7 &   \textbf{7.6$\pm$11.2} \\
\bottomrule

\end{tabular}

}
\end{table*}

\subsubsection{Results in \hqwmca dataset}
The \hqwmca dataset is consists of more challenging attacks as compared to \wmca. Specifically, there are different types of facial tattoos and partial attacks which occupy only a part of the face. These attacks are much harder to detect when they are not seen in the training set, as they appear very similar to bonafide samples.
The experimental results in \hqwmca are tabulated in Table \ref{tab:unseen_hqwmca}. Similar, to \wmca database, the baselines \textit{MCCNN-OCCL-GMM} and \textit{MC-ResNetDLAS} does not perform well in the LOO protocols of \hqwmca database. In addition, the \textit{MC-PixBiS} method, which achieved reasonable performance in \wmca performs poorly \hqwmca dataset. This could be due to the challenging nature of the attacks in the database. It can be seen that the new multi-head architecture proposed, \textit{RGBD-MH-BCE}, already improves the results as compared to all the baselines with an average ACER of $13.3\pm16.5$. Further, with the addition of the \CMFL loss, the ACER further improves to $11.6 \pm 14.8 \%$. The results indicate that the proposed architecture already improves the performance in challenging attacks, and the proposed loss further improves the results achieving state of the art results in the \hqwmca dataset.

\begin{table*}[ht]
\centering
\caption{Performance of the baseline systems and the proposed method in \textbf{unseen} protocols of \hqwmca dataset. The values reported are obtained with a threshold computed for BPCER 1\% in $dev$ set.}
\label{tab:unseen_hqwmca}

\resizebox{0.95\textwidth}{!}
{

\begin{tabular}{lrrrrrrrr>{\columncolor[gray]{0.8}}r}
\toprule
{}                                         &  Flexiblemask &  Glasses &  Makeup &  Mannequin &  Papermask &  Rigidmask &  Tattoo &  Replay &     Mean$\pm$Std  \\
\midrule
MC-PixBiS \cite{george2019deep}                                  &          29.9 &     49.9 &    29.4 &        0.1 &        0.0 &       32.5 &     5.7 &     9.6 &  19.6$\pm$17.1 \\
MCCNN-OCCL-GMM \cite{george_mcocm_tifs2020}                            &          14.2 &     32.7 &    22.0 &        1.5 &        7.1 &       33.7 &     4.2 &    36.6 &  19.0$\pm$13.2 \\
MC-ResNetDLAS \cite{parkin2019recognizing}                              &          23.5 &     50.0 &    33.8 &        1.0 &        2.6 &       31.0 &     5.7 &    15.5 &  20.3$\pm$16.2 \\ \hline
RGBD-MH-BCE            &16.7 &     38.1 &    43.3 &        0.4 &        1.3 &     3.0 &        2.0 &     2.3 &  13.3$\pm$16.5 \\
\textbf{Proposed} &14.8 &     37.4 &    34.9 &        0.0 &        0.4 &        2.4 &     2.4 &     1.0 &  \textbf{11.6$\pm$14.8} \\

\bottomrule
\end{tabular}
}
\end{table*}

To summarize, the newly proposed multi-head architecture itself improves the performance over other baselines. The addition of the \CMFL loss further improves the performance in both \wmca and \hqwmca datasets.

\subsection{Ablation studies}

To further analyze the performance, we conduct various ablation studies using the proposed approach. We carry out these experiments in the \hqwmca dataset as it contains more challenging attacks.

\subsubsection{Effect of $\gamma$} First, we conduct experiments with different values of $\gamma$, we report the mean (value) of all the attacks in the \hqwmca dataset for this comparison. The results with different values of $\gamma$ are shown in Table. \ref{tab:ablation_gamma_hqwmca}. It can be seen that the optimal value of $\gamma = 3$. The model with $\gamma=0$ is equivalent to BCE loss on each of the branches. 

\begin{table}[H]

\centering
\caption{ Ablation study on varying values of $\gamma$ in \hqwmca dataset.}
\label{tab:ablation_gamma_hqwmca}

\begin{tabular}{l>{\columncolor[gray]{0.8}}r}
\toprule
$\gamma$--Values & Mean$\pm$Std \\
\midrule
$\gamma=0$& 13.3$\pm$16.5 \\
$\gamma=1$& 12.5$\pm$15.5 \\
$\gamma=2$& 13.5$\pm$16.2 \\
$\gamma=3$& \textbf{11.6$\pm$14.8} \\
$\gamma=4$& 12.6$\pm$15.2 \\
\bottomrule
\end{tabular}

\end{table}

\subsubsection{Performance with missing channels} Secondly, we evaluate the performance of the models when evaluated with only a single channel at test time. Consider a scenario where the model was trained with RGB and depth, and at the test time, only one of the channels is available. We compare with the mean performance in the \hqwmca dataset, with RGB and depth alone at test time. The results are shown in Table \ref{tab:ablation_individual}. For the baseline \textit{RGBD-MH-BCE}, using RGB alone at test time the error rate is $15.4\pm16.1$, whereas for the proposed approach it improves to $12.0\pm13.9$. The performance improves for the depth channel as well. 

From Table \ref{tab:ablation_individual}, it can be clearly seen that the performance improves, as compared to using BCE even when using a single channel at the time of deployment. This shows that the performance of the system improves when the loss contributions of samples which are not possible to classify by that modality are reduced. Forcing the individual networks to learn a decision boundary leads to overfitting resulting in poor generalization. In the proposed approach, the network can learn a robust classifier for individual channels as well. This observation opens up the possibility to use multi-channel datasets at training time and to deploy the model for the RGB channel only. This could be of practical importance in scenarios where using additional hardware at deployment time is prohibitive, requiring PAD systems to be deployed using legacy RGB cameras. 

\begin{table}[H]

\centering
\caption{ Ablation study using only one channel at deployment time.}
\label{tab:ablation_individual}

\begin{tabular}{ccc}
\toprule
{}         &RGB                     & Depth \\
\midrule

RGBD-MH-BCE  & 15.4$\pm$16.1           & 34.2$\pm$11.6  \\
Proposed  &\textbf{12.0$\pm$13.9}      &\textbf{30.6$\pm$17.5}   \\
\bottomrule
\end{tabular}

\end{table}

\subsubsection{Detailed analysis of the scores}
\begin{figure}[ht!]
\centering
        \includegraphics[width=0.99\linewidth]{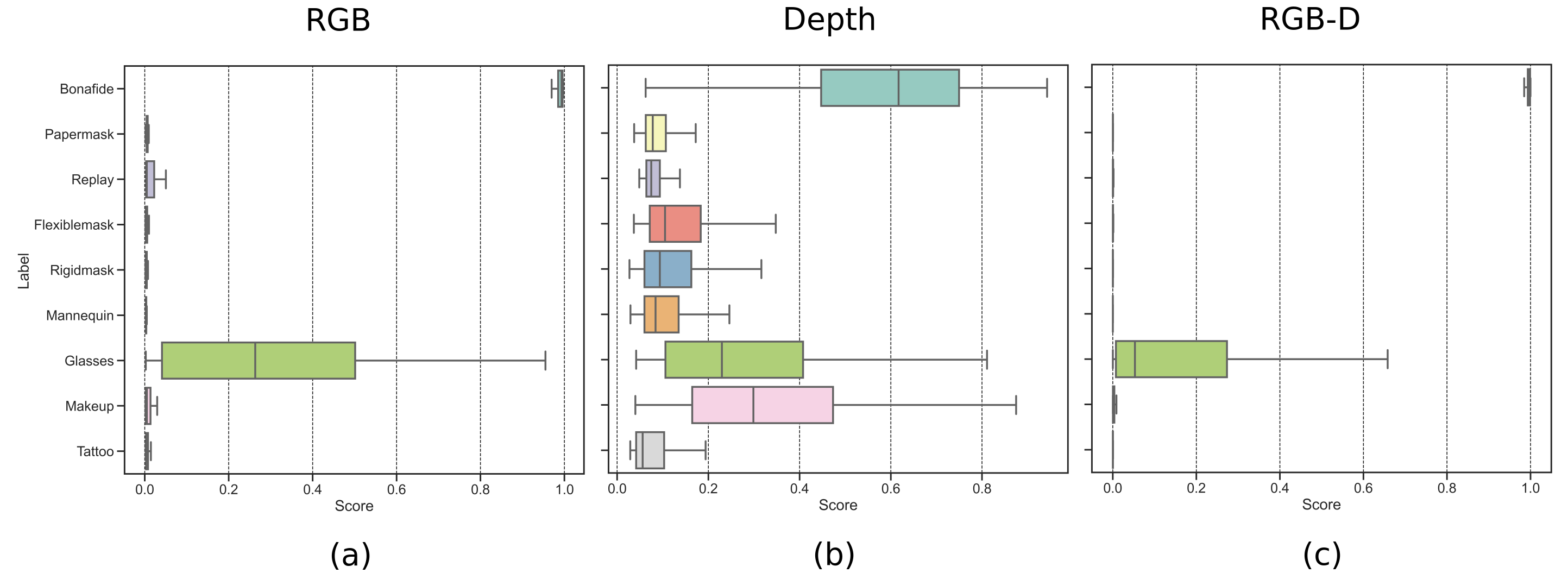}
        \caption{Score distribution for a) RGB, b) Depth, and c) RGB-D branches in the \textit{Proposed} model (Lables: bonafide=1, attacks=0).}

\label{fig:score_gamma3}
\end{figure} 

Here we analyze the score distributions of the proposed model in the $grandtest-c$ protocol of the \hqwmca dataset. As opposed to LOO protocols, the $grandtest-c$ consists of different types of attacks roughly equally distributed in the $train$, $development$, and $evaluation folds$. The multi-head model provides three different scores from each of the branches, i.e., RGB, depth, and RGB-D branches. Here we show the distribution of scores for different types of attacks in the $evaluation$ set of the \hqwmca dataset.

From Fig. \ref{fig:score_gamma3}, it can be seen that the attacks are more compact in RGB head compared to that of depth channel. In depth channel, where classifying attacks are harder, the bonafide distribution is shifted to the left. This can also be seen in the score distribution of attacks in depth channel, i.e., while attack scores are not as low as we desire, still, it is far from bonafide, the loss does not push the depth branch to classify all the attacks correctly, and it performs well on attacks where depth channel contains discriminatory information. Looking at the final `RGB-D' part, it can be seen the proposed loss function makes the classifications using the joint representations from individual channels robust. To summarize, the proposed framework encourages the individual branches to produce non-confident scores for attacks that are confidently classified by the other branch. The discriminative representation from the joint branch, learned in this fashion, results in a robust PAD system.

\subsubsection{Cross database evaluation}

To evaluate the robustness in cross-database scenarios we have performed cross-database experiments between models trained in \wmca and \hqwmca. This evaluation amounts to both cross sensor and unseen attack at the same time which is much more challenging than typical cross-database evaluations. \wmca database uses the Intel Realsense SR300 camera which returns both RGB and depth streams. IntelRealsense SR300 uses dot pattern projection for depth computation. Whereas in \hqwmca the RGB channel is acquired using a high-quality Basler acA1921-150uc camera and the depth image was captured with Intel RealSense D415 (which uses stereo for computing depth). The mismatch between the sensors and quality between the sensors used in these datasets makes the performance degrade. The results from the cross-dataset test are tabulated in Table. \ref{tab:cross_test_eer}. The models were trained on the \textit{grandtest} protocol of each dataset and was evaluated on the $dev$ and $eval$ set of the other dataset. We report both the intra-and cross-database performance for the sake of completeness. It can be seen that in the intra dataset evaluations, the proposed method achieves good performance in both the datasets. However, in the cross-database evaluations, the performance degrades. The mismatch between the sensors could be one reason for the degradation in performance. Performance is slightly better when the model trained in \hqwmca is evaluated on \wmca which could be due to the wider variety of attacks present in the training set. Moreover, it can be seen that methods with better performance in the target dataset, perform worse in the source dataset.

\begin{table}[ht]
\begin{center}
\footnotesize
\caption{The results from the cross-database testing between \wmca and \hqwmca datasets. HTER (\%) values computed in $eval$ set for threshold computed in $dev$ set using EER criteria are reported in the table.}
\label{tab:cross_test_eer}

\resizebox{0.95\columnwidth}{!}{%
\begin{tabular}{@{}c|c|>{\columncolor[gray]{0.8}}c|c|>{\columncolor[gray]{0.8}}c@{}}
\toprule
\multirow{2}{*}{Method} & \multicolumn{2}{c|}{\begin{tabular}[c]{@{}c@{}}trained on\\ \wmca \end{tabular}}                                     & \multicolumn{2}{c}{\begin{tabular}[c]{@{}c@{}}trained on\\ \hqwmca \end{tabular}}                                       \\ \cline{2-5}  
                        & \begin{tabular}[c]{@{}c@{}}tested on\\  \wmca \end{tabular} & \begin{tabular}[c]{@{}c@{}}tested on\\ \hqwmca \end{tabular} & \begin{tabular}[c]{@{}c@{}}tested on\\ \hqwmca \end{tabular} & \begin{tabular}[c]{@{}c@{}}tested on \\ \wmca \end{tabular} \\ \midrule

MC-PixBiS                         &1.8     &25.0      &3.0      &11.7    \\ 
MC-ResNetDLAS                     &4.2     &34.1      &8.2      &21.7    \\ 
MCCNN-OCCL-GMM                    &3.3     &23.1      &16.4     &8.5   \\ \hline

\textbf{Proposed}                 &1.7     &29.1      &2.6      &18.2  \\
 \bottomrule
\end{tabular}
}
\end{center}

\end{table}

\subsection{Discussions}

From the experiments in \wmca and \hqwmca, it can be seen that the proposed approach outperforms the state of the art methods. We have also shown that the proposed approach is useful in cases where multiple channels are available at training time and when there is a need to deploy with individual channels.  Even though the proposed approach works well in intra dataset scenario, cross-database performance needs further improvement. As future work, more preprocessing and data augmentation strategies could be developed to alleviate the discrepancies between datasets to improve the cross-database performance. While we have selected RGB and depth channel for this study, mainly due to the availability of off the shelf devices consisting of these channels, it is trivial to extend this study to other combinations of channels as well, for instance, RGB-Infrared, and RGB-Thermal.

\section{Conclusions}

In this work, we proposed a new architecture for RGB-D presentation attack detection, applicable to other multi-channel classification problems as well. Also, we proposed a new cross-modal focal loss function which could be useful for two-stream networks. The proposed cross-modal focal loss function modulates the loss contribution of samples based on the confidence of individual channels. The proposed framework can be trivially extended to multiple channels and different classification problems where information from one channel alone is inadequate for classification. This loss forces the network to learn complementary, discriminative, and robust representations for the component channels. The structure of the framework makes it possible to train models using all the available channels and to deploy with a subset of channels. Extensive evaluations in two publicly available datasets demonstrate the effectiveness of the proposed approach.

\section*{Acknowledgment}

Part of this research is based upon work supported by the Office of the
Director of National Intelligence (ODNI), Intelligence Advanced Research
Projects Activity (IARPA), via IARPA R\&D Contract No. 2017-17020200005.
The views and conclusions contained herein are those of the authors and
should not be interpreted as necessarily representing the official
policies or endorsements, either expressed or implied, of the ODNI,
IARPA, or the U.S. Government. The U.S. Government is authorized to
reproduce and distribute reprints for Governmental purposes
notwithstanding any copyright annotation thereon.


{\small
\bibliographystyle{ieee}
\bibliography{egbib}
}
\end{document}